\PassOptionsToPackage{dvipsnames}{xcolor}
\documentclass[sigconf,nonacm]{acmart}
\AtBeginDocument{%
  \providecommand\BibTeX{{%
    \normalfont B\kern-0.5em{\scshape i\kern-0.25em b}\kern-0.8em\TeX}}}

\let\oldfbox\fbox
\renewcommand{\fbox}[1]{{\color{gray}\oldfbox{\color{black}#1}}}

\usepackage{xspace}
\usepackage{xcolor}
\usepackage{hyperref}
\usepackage{tikz}
\usepackage{multirow}

\usepackage{enumitem}
\usepackage{array}
\newcommand{\chip}[2]{%
  \tikz[baseline=(text.base)]{%
    \node[fill=#1!15, rounded corners=2pt, inner xsep=1.5pt, inner ysep=0.2pt, text=#1] (text) {\strut #2};%
  }%
}

\definecolor{PaperColor}{HTML}{57534e}
\definecolor{PapersColor}{HTML}{15803d}
\definecolor{ReportColor}{HTML}{7e22ce}

\newcommand{\sys}{Ream\xspace}
\newcommand{\paper}{\chip{PaperColor}{\texttt{.paper}}\xspace}
\newcommand{\papers}[1][]{\chip{PapersColor}{\texttt{#1.papers}}\xspace}
\newcommand{\report}[1][]{\chip{ReportColor}{\texttt{#1.report}}\xspace}

\newcommand{\explorations}{Agent Explorations\xspace}

\begin{document}

\title{Ream: Unfolding Mutual Awareness in Human-Agent Workspaces}

\author{Peiling Jiang}
\authornote{This work was completed during an internship at the Allen Institute for AI (Ai2).}
\email{peiling@ucsd.edu}
\orcid{0000-0003-4447-0111}
\affiliation{%
  \institution{University of California San Diego}
  \city{La Jolla}
  \state{California}
  \country{USA}
}

\author{Sangho Suh}
\email{sanghos@allenai.org}
\orcid{0000-0003-4617-5116}
\affiliation{%
  \institution{Allen Institute for AI}
  \city{Seattle}
  \state{WA}
  \country{USA}
}

\author{Varsha Kishore}
\email{varshak@allenai.org}
\affiliation{%
  \institution{Allen Institute for AI}
  \city{Seattle}
  \state{WA}
  \country{USA}
}

\author{Jonathan Bragg}
\email{jbragg@allenai.org}
\orcid{0000-0001-5460-9047}
\affiliation{%
  \institution{Allen Institute for AI}
  \city{Seattle}
  \state{WA}
  \country{USA}
}

\author{Haijun Xia}
\email{haijunxia@ucsd.edu}
\orcid{0000-0002-9425-0881}
\affiliation{%
  \institution{University of California San Diego}
  \city{La Jolla}
  \state{California}
  \country{USA}
}

\author{Pao Siangliulue}
\email{paos@allenai.org}
\orcid{0009-0006-8042-885X}
\affiliation{%
  \institution{Allen Institute for AI}
  \city{Seattle}
  \state{WA}
  \country{USA}
}

\author{Daniel S. Weld}
\email{danw@allenai.org}
\orcid{0000-0002-3255-0109}
\affiliation{%
  \institution{Allen Institute for AI}
  \city{Seattle}
  \state{WA}
  \country{USA}
}

\author{Amy X. Zhang}
\email{axz@cs.uw.edu}
\orcid{0000-0001-9462-9835}
\affiliation{%
  \institution{University of Washington}
  \city{Seattle}
  \state{Washington}
  \country{USA}
}

\author{Joseph Chee Chang}
\email{josephc@allenai.org}
\orcid{0000-0002-0798-4351}
\affiliation{%
  \institution{Allen Institute for AI}
  \city{Seattle}
  \state{WA}
  \country{USA}
}

\renewcommand{\shortauthors}{Jiang et al.}

\begin{abstract}
  As AI agents work alongside humans in shared workspaces, a mutual awareness challenge arises: agents act at speeds that outpace human monitoring, and users' evolving interests are not always expressed in chat. This challenge is especially pressing in literature review, where both parties retrieve, read, and synthesize a growing body of papers. We present \mbox{\sys}, a literature review workspace that supports mutual awareness through structured artifacts, bidirectional engagement tracking, and localized visualizations. Users can see each party's activity within these documents, and agents can retrieve the same history to guide their work. In studies with eighteen researchers, participants used these traces to inspect evidence, steer agents, communicate through annotations, and reflect on their research focus. Shared histories also helped agents build on earlier work. These findings inform how engagement traces within shared documents can support transparency, personalized assistance, and coordination in human-agent knowledge work.

\end{abstract}



\begin{teaserfigure}
  \centering\includegraphics[width=\textwidth]{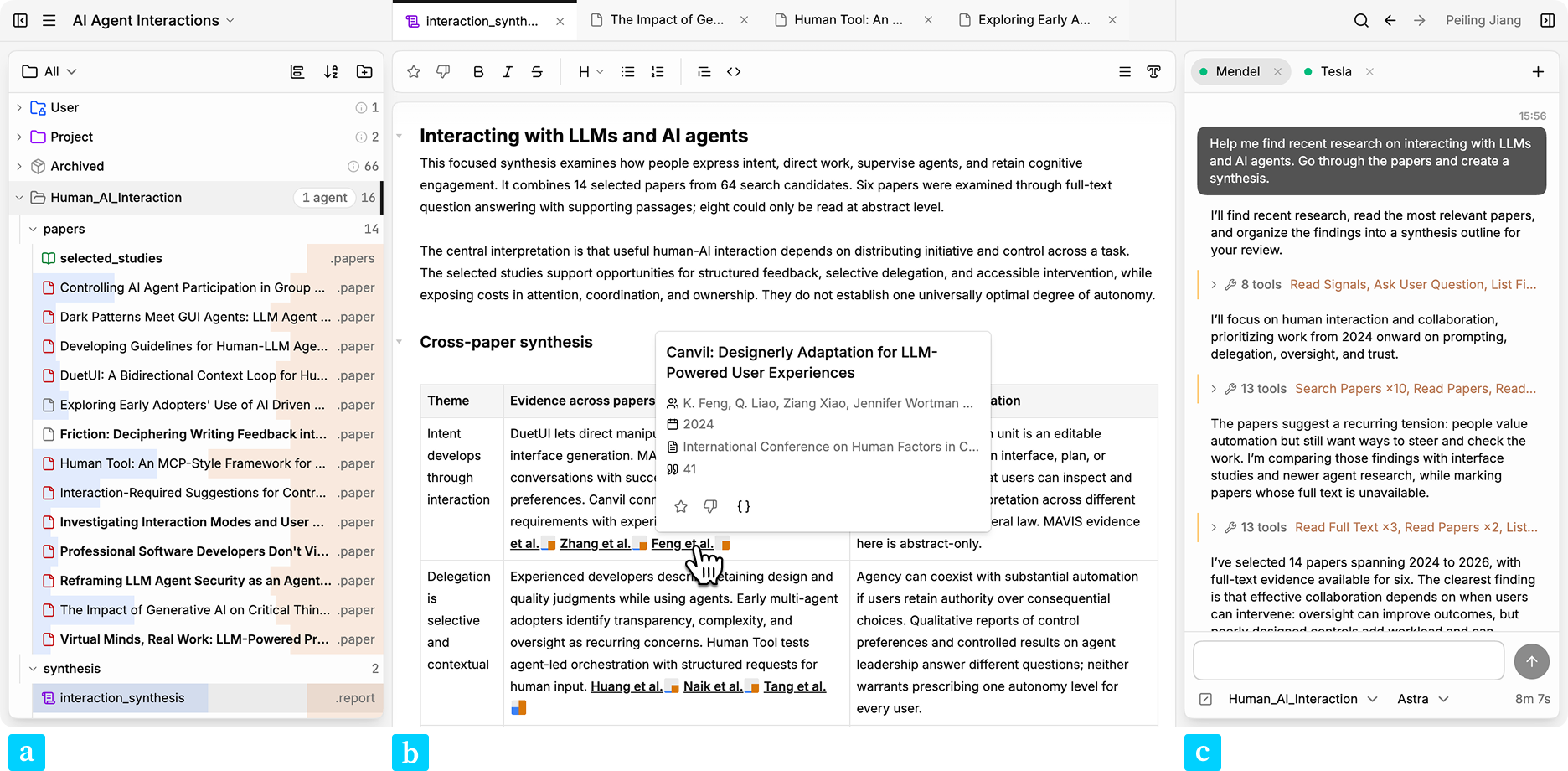}
  \caption{\sys is a human-agent collaborative literature review workspace, tracking and visualizing both the user's and agents' actions to facilitate mutual awareness. The file tree (a) shows different amounts of activity on each file, with blue indicating user activity and orange indicating the agents'. The tabbed file viewer (b) shows a \report file with paper citation chips. The right chat panel (c) allows the user to create and instruct multiple agents across workspace folders at the same time.}
  \Description{The Ream workspace has three panels: a file tree with blue user activity bars and orange agent activity bars, a tabbed report viewer with interactive citations and a paper preview card, and a chat panel for coordinating agents.}
  \label{fig:overview}
\end{teaserfigure}

\maketitle

\section{Introduction}

The interaction paradigm between humans and AI is expanding from direct manipulation to multi-turn conversations and then to deeper, more prolonged collaboration in shared workspaces with many artifacts~\cite{shao2025future,feng2025levels,CowPilot,feng2024cocoa,Petridis2026CompassVR}.
Most notably, software developers are rapidly adopting coding agents---such as Claude Code~\cite{ClaudeCode} and Codex~\cite{Codex}---that can autonomously navigate large codebases, edit files, and run tests by interleaving reasoning with action and tool use~\cite{yao2022react,schick2023toolformer}.
Users can interact with agents through chat (or CLIs) with simple pointing mechanisms---referencing a file or a line of code in conversation~\cite{Kirk2007TurnIT}.
Since such agents can flexibly adopt different non-coding ``skills,'' such as calling APIs for document retrieval or visual design critique, users are also increasingly deploying AI agents across workspaces in other domains, such as design~\cite{Lu2024AIAF, son2026hand}, ideation~\cite{Shaer2024Brainwriting}, writing~\cite{Lee2024Writing,Lee2022CoAuthorDA}, and academic research~\cite{gao2025democratizingaiscientistsusing, gao2025txagent} as active collaborators.

Meanwhile, decades of HCI research have explored interfaces and interactions that facilitate real-time human-human collaboration across scenarios such as exploratory search~\cite{morris2007search}, writing~\cite{larsen2019territorial}, coding~\cite{goldman2011real}, and ideation~\cite{Geyer2012IdeaVisAH,Clayphan2011FirestormAB}. These works point to the importance of \emph{mutual awareness}, the peripheral understanding of each other's activities, to facilitate real-time task coordination, build coherent mental models from inspecting past actions, and avoid conflicts in the shared workspace~\cite{gutwin2002descriptive,dourish1992awareness,territorial}.
However, while the importance of mutual awareness between human collaborators is well established and studied, little work has examined designs to facilitate mutual awareness between humans and their AI agents.
As the paradigm continues to shift towards human-agent collaboration in shared workspaces, it becomes important to explore how these principles apply~\cite{Lehmann2025CollaborativeDE, Shen2025BidirectionalHA}.
More specifically, we build on Heer's argument that shared representations through which people and algorithms can both contribute to a task provide a basis for combining computational assistance with human control~\cite{heer2019agency}.

Existing interfaces offer limited support for mutual awareness between users and agents. For example, chat interfaces present an agent's reasoning traces and actions in chronological order, but as agents move into shared workspaces, they examine, create, and modify files at speeds that outpace user monitoring from the linear chat view. As a result, it can be prohibitively difficult for users to map the long, linear reasoning and action traces to the many files the agent has affected in the hierarchical folder structure.
Likewise, agents typically lack awareness of how the user has been engaging with the workspace over time---which files they read repeatedly, which they deprioritized---preventing them from proactively aligning with the user's evolving goals.
More fundamentally, files in a project capture only their current state, but not how they came to be. For example, they do not record which sources an agent retrieved to support their creation or modification, or whether the user has recently visited, repeatedly edited, or ignored them. Keeping a history of these actions alongside the documents would help \emph{users} better follow the agent's work in the workspace and allow \emph{agents} to learn about the user's focus and preferences.

In this paper, we explore how mutual awareness can be integrated into human-agent workspaces to help both parties track, interpret, and adapt to each other's actions.
We chose literature review as our domain to explore awareness unfolding in human-agent collaboration---a representative process of knowledge work and sensemaking that involves a large volume of heterogeneous materials, evolving research questions, and iterative cycles of searching, reading, organizing, and synthesizing~\cite{russell1993cost,pirolli2005sensemaking,Synergi,Liao2024Research}.

Literature review is particularly well suited to our exploration for several reasons. Researchers need to assess claims against sources while developing questions and interpretations that can change during exploration. They often \emph{desire} to engage directly with papers even when an AI agent can retrieve and summarize them~\cite{Haddad2026Understanding,Relatedly,Sultanum2020Understanding}.
Engagement extends beyond chat---researchers revisit, annotate, and edit documents, while agents leave traces through searches, questions, and extracted passages. As the body of documents grows, making these signals mutually accessible becomes not merely helpful but essential for agents to follow researchers' evolving preferences and for researchers to inspect and steer agent work.

To explore this gap, we introduce \sys, a shared, human-agent workspace for literature review that tracks and visualizes the actions of both the user and agents as they work across documents.
\sys represents the literature review workflow as structured, file-based artifacts---individual papers, curated collections, and synthesis reports---that both parties can inspect and act on.
Engagement signals (such as opens, edits, annotations, and revisits) accumulate over time and appear in localized visualizations in the file tree and documents, helping users maintain awareness of where agents have focused their effort while allowing agents to leverage the user's footprint to proactively align their assistance.

Our work makes the following contributions:
\begin{enumerate}
    \item \textbf{A design approach for artifact-centered mutual awareness.} We adapt workspace-awareness principles to human-agent knowledge work: track both parties' interactions with shared artifacts over time, visualize these interactions within the documents, and make them available to agents.
    \item \textbf{\sys}---a literature review workspace that implements this approach through structured artifacts, engagement tracking, localized visualizations, and agent retrieval.
    \item \textbf{Two user studies} (\mbox{lab $N = 12$ and longitudinal $N = 6$}) showing how researchers used traces to inspect evidence, steer assistance through workspace activity, reflect on their own focus, and carry context across multiple agents. These patterns recurred in longitudinal use, informing design implications for ongoing human-agent collaboration.
\end{enumerate}

\section{Background and Related Work}

Our work builds on research on awareness in human-human and human-agent collaborative workspaces and extends these principles through explorations of emerging conversational coding agents and AI literature review tools.

\subsection{Awareness in Human-Human Collaborative~Workspaces}

We take inspiration from the rich prior work in CSCW on how awareness supports effective collaboration between people to inform our design of shared workspaces for human-agent collaboration.
These studies explored mutual awareness as a way to help collaborators maintain passive, peripheral awareness of one another's activity, contextualize one's own work, and act in ways complementary to the other's contributions~\cite{dourish1992awareness,gutwin2002descriptive,clark1991grounding}.
Gutwin and Greenberg operationalized the definition into the widely adopted descriptive framework for workspace awareness in real-time groupware, decomposing awareness into elements organized around \emph{who} is working on \emph{what}, \emph{where} they have been in the shared workspace, and what specific \emph{actions} they have been taking~\cite{gutwin2002descriptive,gutwin1996workspace,gutwin1998design}.

Prior work explored leveraging interaction traces and usage history on shared artifacts to establish and communicate awareness~\cite{hill1992wear,gutwin1996usability}.
Hill et al.\ introduced ``computational wear'' to visualize read and edit activities in documents, drawing an analogy to how physical objects naturally accumulate visible traces of use~\cite{hill1992wear}.
Clark and Brennan showed how collaborators establish common ground through incremental evidence of mutual understanding~\cite{clark1991grounding}.
Gergle et al.\ showed that shared visual information can serve as both an awareness channel and a grounding resource, reducing the need for verbal coordination~\cite{gergle2013using}.
In collaborative search, SearchTogether~\cite{morris2007search} and CoSearch~\cite{amershi2008cosearch} let partners share query histories and divide search results.
In collaborative writing, Larsen-Ledet and Korsgaard studied how co-authors navigate ownership and awareness of each other's edits in a shared document~\cite{larsen2019territorial}.
In collaborative ideation, Firestorm supported brainstorming on shared, spatial surfaces with visibility of everyone's contributions~\cite{Clayphan2011FirestormAB}.

Across these domains, collaborators' activities are captured as visible, persistent traces in shared artifacts, enabling coordination without explicit communication~\cite{gutwin1998design}.
\sys adopts this principle at three levels: it represents the literature review workflow through shared, structured artifacts, giving both users and agents concrete materials to act on and a shared basis for visualizing past actions for awareness; it then applies localized visualizations tailored to each document type in the workspace; and it tracks interaction history for both users and agents to surface attention and contributions.

\subsection{Awareness in Human-Agent Collaborative~Workspaces}

A growing body of research argues that AI should be designed not as a standalone replacement for humans but as a collaborator, echoing the ``Man-Computer Symbiosis'' vision~\cite{licklider1960man,Engelbart1962AugmentingHI,horvitz1999principles}.

Recently, Haupt and Brynjolfsson argued that the machine learning community should shift from evaluating AI systems in isolation toward ``centaur evaluations'' in which humans and AI jointly solve tasks~\cite{haupt2025}. Feng et al.\ further formalized this spectrum through five levels of agent autonomy defined by the role the user assumes---operator, collaborator, consultant, approver, or observer. They argue that autonomy should be a deliberate design decision rather than an inevitable consequence of increasing capability, establishing the basis for research on human-AI collaborative workspaces and mechanisms essential to effective collaboration~\cite{feng2025levels}.

For such collaboration to work, both parties need to know what the other is doing and why~\cite{lai2021towards,amershi2019,he2023interaction}.
Zhang et al.\ found that when the agent modeled its partner's goals and beliefs, humans felt more understood---and that implicit coordination through workspace actions was as effective as verbal communication~\cite{zhang2024mutual}.
Chen et al.\ proposed a three-level model of agent transparency: communicating what the agent is doing, why, and what it will do next~\cite{chen2018situation}.
Bansal et al.\ found that human-AI team performance hinges on whether the human knows when the AI is likely to be wrong, not just how accurate it is overall~\cite{bansal2019beyond}.
Son et al.\ showed that in design workspaces, making agent execution visible step-by-step lets users anticipate actions, intervene, and coordinate parallel work~\cite{son2026hand}.
\mbox{\sys} extends these principles by tracking how users and agents work with the same documents over time. Users can see which papers an agent has retrieved, what questions it has asked, and which passages it has extracted. Agents can similarly retrieve the user's interaction history to identify papers they revisited or notes they added, helping both parties follow each other's work beyond the conversation.

This accumulated awareness also informs when and how the agent should take initiative.
Prior work has explored making agents more proactive in conversation---through post-training for better clarification and follow-up suggestions~\cite{wu2025collabllm}, prompting techniques that simulate ``inner thoughts'' for knowing when to chime in~\cite{liu2024proactive}, building generalized user models~\cite{shaikh2025gum}, and broader efforts on proactive dialogue in the LLM era~\cite{liao2023proactive,deng2023prompting,deng2023rethinking}.
While much prior work considers the ongoing conversation to identify opportunities for proactivity, \sys takes a different approach: because it already tracks how attention accumulates across the workspace over time, it can leverage these signals---which papers the user reads closely, what notes they take, which summaries they revisit---to infer evolving interests and proactively suggest tasks aligned with what the user cares most about.

\subsection{AI-Supported Literature Review Tools}

AI-powered tools for scientific literature review have advanced rapidly in recent years, supporting nearly every stage of the literature review workflow~\cite{Sultanum2020Understanding}, from searching and screening~\cite{van2021open,ouzzani2016rayyan,khraisha2024can,cassell2025analysis,spillias2024human,paperfinder,Threddy}, to reading and comprehension~\cite{lo2024semantic,head2021augmenting,fok2023scim,chang2023citesee,Threddy,ma2025garden}, and finally creating comprehensive syntheses~\cite{wang2024autosurvey,martin2024shallow,alshami2023harnessing}.
Academic tools such as the open-source Ai2 Scholar QA~\cite{scholarqa} and commercial tools such as Elicit~\cite{elicit} and Perplexity~\cite{perplexity} now combine many of these capabilities, retrieving hundreds of papers, thematically clustering quotes, and synthesizing attributed multi-section reports, all mediated through a query-answering paradigm.

Haddad et al.\ analyzed over 200,000 queries from Ai2 Scholar QA~\cite{scholarqa} and Ai2 Paper Finder~\cite{paperfinder} and found that users submit significantly longer, more complex queries than in traditional search engines---pasting draft paragraphs or meeting notes as context, delegating complex, multi-stage tasks beyond question answering---suggesting users increasingly seek AI not as a reactive tool but as a collaborative partner~\cite{Haddad2026Understanding,shao2025future,huang2025control}.
Current deep-research-based tools, however, mostly generate a single report for each user query~\cite{xu2025comprehensive,scholarqa}. They cannot adapt to flexible user requests and delegations across the full review workflow. Nor do they provide a shared, evolving information space for users to inspect and steer the AI literature review process~\cite{chen2026singleshotwritingdeepresearch}.

\mbox{\sys} embeds AI agents within a shared workspace where researchers and agents work with the same evolving papers, curated collections, and synthesis reports. Researchers can directly inspect intermediate outputs and flexibly steer the agent as the review develops. Building on systems such as CiteSee, which uses a reader's interaction history to contextualize citations~\cite{chang2023citesee}, \mbox{\sys} tracks interactions by both users and agents. These interactions are visualized within the documents for users and made retrievable by agents, helping both parties maintain mutual awareness throughout the review process.

\section{Design Goals}

We use mutual awareness to describe how users and agents keep track of each other's activities in the shared workspace and adapt their work accordingly.
As collaboration with AI agents moves from chat into shared workspaces, \emph{what does mutual awareness look like here, and how do we design for it?}
The challenge has two dimensions for both parties: \emph{where} both parties' work is distributed across a growing set of documents, and how both parties' attention shifts \emph{over time}. We address this through three design goals:

\begin{itemize}[leftmargin=2.5em]

\item [\textbf{DG1}]
\textbf{Represent knowledge workflows as domain-specific artifacts across shared workspaces.}
Collaborative literature review requires both parties to inspect and build on each other's intermediate work---search results, reading notes, paper collections, and syntheses~\cite{Sultanum2020Understanding,Haddad2026Understanding}.
Awareness needs to be grounded in concrete, shared artifacts.
Therefore, \sys provides structured artifacts representing papers, paper collections, and synthesis reports to materialize the literature review workflow for both the user and agents to review and act on, providing surfaces on which awareness can then be visualized~\cite{Threddy,Sultanum2020Understanding}.

\item [\textbf{DG2}]
\textbf{Track engagement over time to establish mutual awareness.}
The current status of the workspace is naturally visible to both the user and the agent~\cite{ClaudeCode,cursor}.
However, a snapshot is insufficient for mutual awareness in literature review workflows:
agents operate at speeds and volumes that outpace human monitoring, so users need to see agents' accumulated effort across documents beyond the current action; similarly, learning the user's evolving interests---which papers they read closely, which they revisit, which they deprioritize or remove---requires more than knowing which files currently exist or are open.
Therefore, \sys tracks interaction signals---opens, edits, stars, notes---over time, building a model of engagement that both parties can draw on~\cite{chang2023citesee}.

\item [\textbf{DG3}]
\textbf{Surface bidirectional attention through localized visualizations and agentic retrieval.}
As interaction records accumulate across shared artifacts, users need to see how they and the agents have engaged with these documents. Embedding effort and attention visualizations in the file tree and documents supports awareness while reading and writing, without switching to separate views that interrupt the workflow~\cite{Synergi,logit}. Correspondingly, \mbox{\sys} makes the same engagement signals available through dedicated query tools, allowing agents to see which papers the user repeatedly revisits or deprioritizes and adapt their retrieval, organization, and synthesis accordingly.

\end{itemize}

Table~\mbox{\ref{tab:awareness}} maps these design choices to the workspace-awareness framework~\mbox{\cite{gutwin2002descriptive}}. Interaction histories show who acted, what they did, which document they worked on, and when.

\begin{table}[t]
\small
\setlength{\tabcolsep}{4pt}
\caption{Workspace-awareness elements adapted to human-agent artifacts, with the questions each mechanism helps collaborators address.}
\label{tab:awareness}
\begin{tabular}{@{}>{\raggedright\arraybackslash}p{0.12\columnwidth}>{\raggedright\arraybackslash}p{0.38\columnwidth}>{\raggedright\arraybackslash}p{0.42\columnwidth}@{}}
\toprule
Element & Awareness question & Ream mechanisms \\
\midrule
Who & Which party engaged with this artifact? & User/agent attribution in bars and histories \\
What & What work has occurred~here? & Search, read, write, chat, and organize events \\
Where & Which sources and passages received attention? & Traces on files, references, and PDF passages \\
When & When was this artifact visited or changed? & Timestamped per-file histories \\
How & How was a source used to address a question? & Exploration questions, answers, and extracted quotes \\
Why & What goals or preferences help explain an action? & Notes, stars/downvotes, and agent questions \\
\bottomrule
\end{tabular}
\end{table}

\section{Agentic Literature Review with \sys}

To explore the design goals, we built \sys,\footnote{\url{https://ream.sh}} a web-based literature review workspace where researchers collaborate with AI agents through a shared set of structured documents.
\sys's interface consists of a file tree panel to view all workspaces and files, a tabbed document viewer and editor at the center, and a chat panel to manage and chat with agents on the right (Figure~\ref{fig:overview}).

\sys's AI agents leverage specialized literature review tools for searching and reading papers, and for creating, editing, and organizing documents in workspaces (Appendix~\ref{sec:implementation}).
Each \sys project can contain multiple workspaces for isolated work.
For example, as an agent fetches papers in an exploratory search workspace, a curated workspace on a different topic remains unaffected.
Multiple agents can run in parallel across the same or different workspaces, supporting concurrent searching and synthesis.

\sys represents each stage of the literature review workflow as a domain-specific file type \textbf{(DG1)}, from papers and collections to synthesis reports. These structured artifacts give both the researcher and the agent concrete surfaces to inspect, annotate, and build upon---and serve as the grounding for awareness tracking and visualization (Section~\ref{sec:visualizations}).

\subsection{Representing the Literature Review Workflow with Structured Artifacts (DG1)}
\label{sec:document-types}

For the researcher and agents to collaborate in literature review, the workspace needs concrete artifacts for both parties to review and act on.
Building on prior literature review systems, we represent individual papers, curated collections, and synthesis reports as three types of structured documents that both parties can inspect and build on~\cite{Sultanum2020Understanding,Threddy,Relatedly}. Tracking their interactions with these documents allows each side to see how the other has contributed to the review process.

\noindent\textbf{Paper documents.}
Information for a paper is typically scattered across publisher webpages, locally stored PDFs, and personal notes and annotations.
This fragmentation makes it difficult to synthesize findings or maintain a coherent picture of each source throughout the review and subsequent revisits.
Therefore, \sys consolidates each paper into a single \paper file containing both authoritative information about the paper---from metadata\footnote{From the Semantic Scholar API, including title, authors, venue, year, abstract, summary, reference and citation counts, links, figures, and tables.} to full text and PDF---and researcher notes, extracted quotes, and annotations specific to the ongoing review process.
This gives both parties a self-contained artifact to read, annotate, and reference.

\begin{figure}
\centering
\includegraphics[width=\columnwidth]{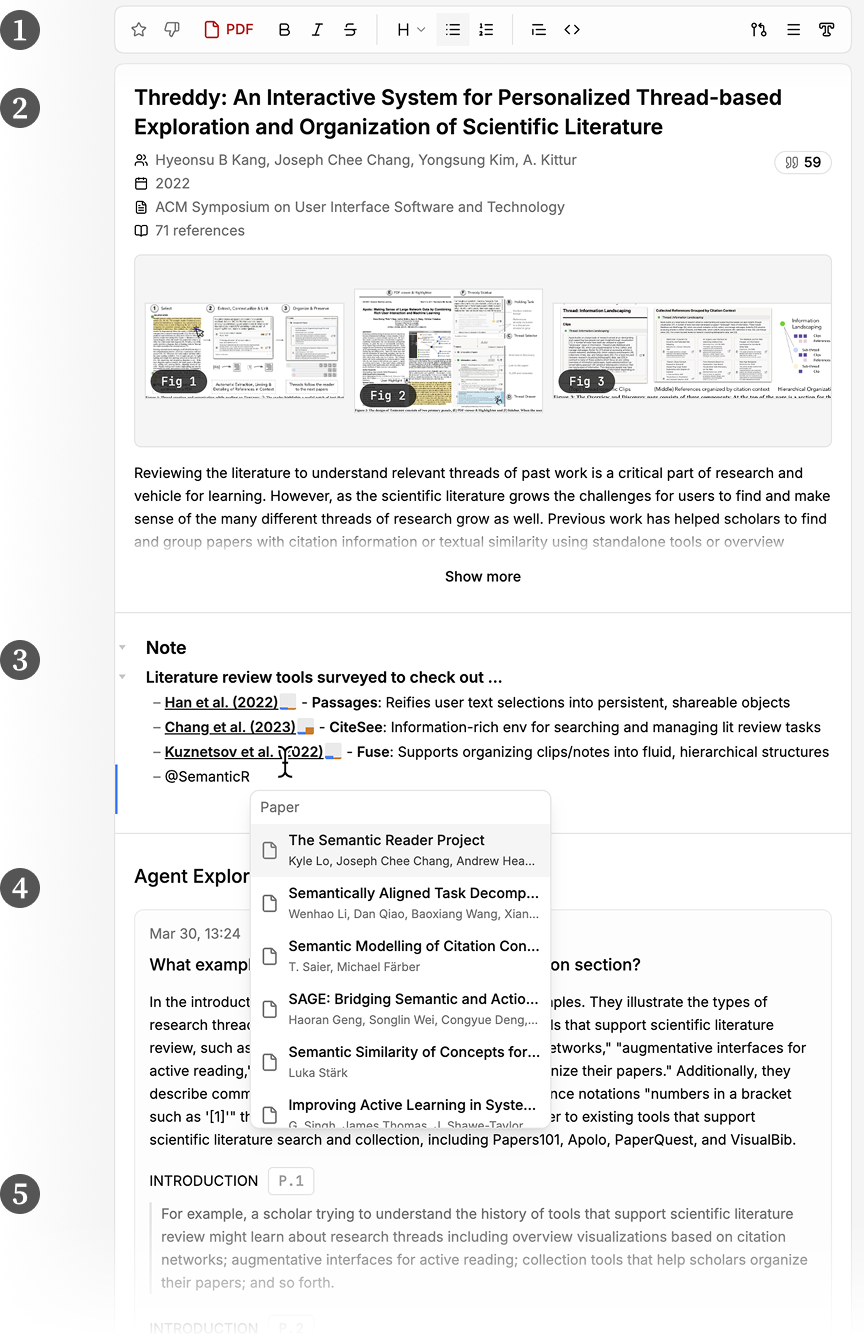}
\caption{The \texttt{.paper} file view includes: (1) controls allowing users to Star or Downvote a paper to indicate their interests; (2) paper metadata and figures; (3) user notes (here, the user took notes about additional papers found while reading, which are in turn used by the agent to generate proactive suggestions); and (4) the Agent Explorations section with questions, answers, and quotes extracted from the paper (5).}
\Description{A paper document with Star and Downvote controls, paper metadata and figures, user notes, and an Agent Explorations section containing questions, answers, and evidence quotes from the paper.}
\label{fig:paper}
\end{figure}

A \paper file may be created by the agent during a search or by parsing a user-uploaded PDF.
The user can add notes using the inline editor (Figure~\ref{fig:paper}).
The agent can use the \texttt{read\_full\_text} tool to ask an LLM to extract information from the full paper and answer a specific question with verbatim evidence quotes (which show up in the PDF as agents' annotations and ``read wear''~\cite{hill1992wear}), or \texttt{read\_papers} to retrieve metadata and all previous explorations.
The user notes and agent explorations make each party's engagement with the paper visible to the other within a single artifact.

While the same paper may appear in multiple workspaces and folders, all \paper files in the file tree that represent the same paper point to a single underlying file, so notes and annotations made in one context are immediately visible everywhere.\newline

\noindent\textbf{Paper collections.}
As a literature review progresses, dozens to hundreds of papers may accumulate from many searches and can easily collapse into an indistinguishable list.
\sys organizes ordered lists of papers as \papers files to provide one more layer of structure and provenance.
A \papers file is automatically created for each paper search, allowing the agent to easily revisit all papers found in a session.
The user and agents can also create collections to curate papers by theme or task for easy revisiting: the user can scan, sort, and filter at the collection level, while the agent can reference an entire collection as a shortcut, e.g., passing \texttt{uist-2026.papers} to \texttt{read\_papers} rather than listing each paper.\newline

\begin{figure}
\centering
\includegraphics[width=\columnwidth]{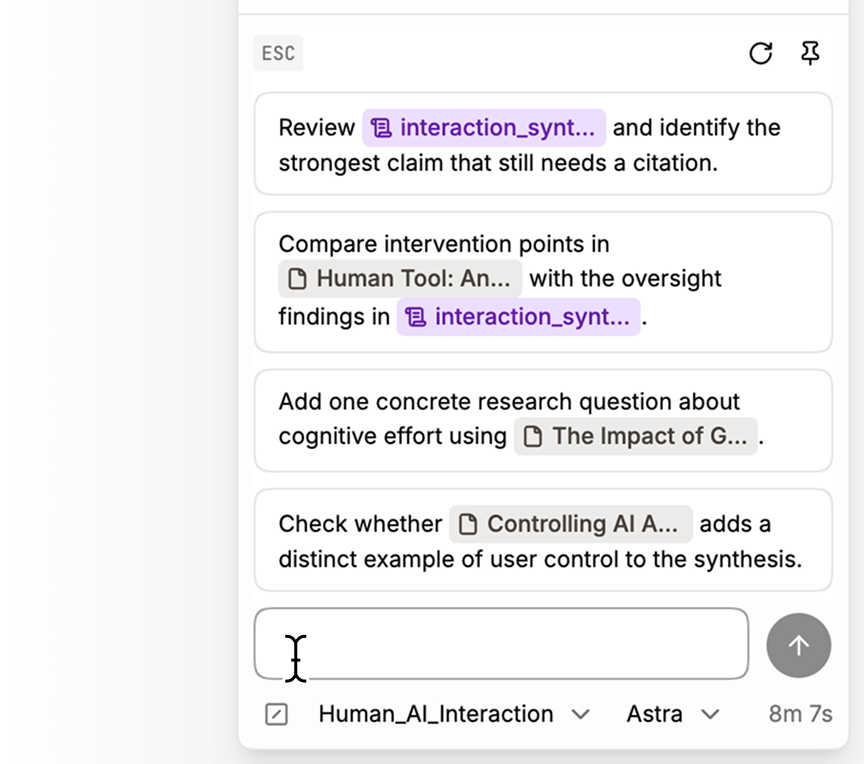}
\caption{When the chat input is focused, \sys proactively generates task suggestions based on user activities.}
\Description{Four task suggestions above the chat input propose reviewing a synthesis for unsupported claims, comparing interaction patterns across papers, adding a research question, and checking whether another paper adds a distinct form of user control.}
\label{fig:suggestion}
\end{figure}

\noindent\textbf{Synthesis reports.}
The primary output of a literature review is a written synthesis that connects themes and claims to sources.
\sys stores syntheses as editable \report files that users and agents can revisit and revise.
Paper references are rendered as clickable links, maintaining live connections to underlying \paper files.

In \sys, the agent by default drafts the \report as an outline with a brief summary of each theme, allowing the user to scan and reshape it before the agent writes in detail. Both parties can then iteratively contribute.
Reports can also include Markdown tables and diagrams to compare papers or illustrate relationships between them, with references linking back to the individual papers~\cite{heer2023living}.
With multiple \report files, the user can also explore different framings of the same body of literature in parallel, without overwriting previous syntheses.

In the file tree, \papers and \report files can act as ``smart folders'': their referenced papers appear as nested items, giving the user a spatial overview of collection membership and a familiar structure for locating papers~\cite{bento,amershi2008cosearch}.

\begin{figure*}
\centering
\includegraphics[width=\textwidth]{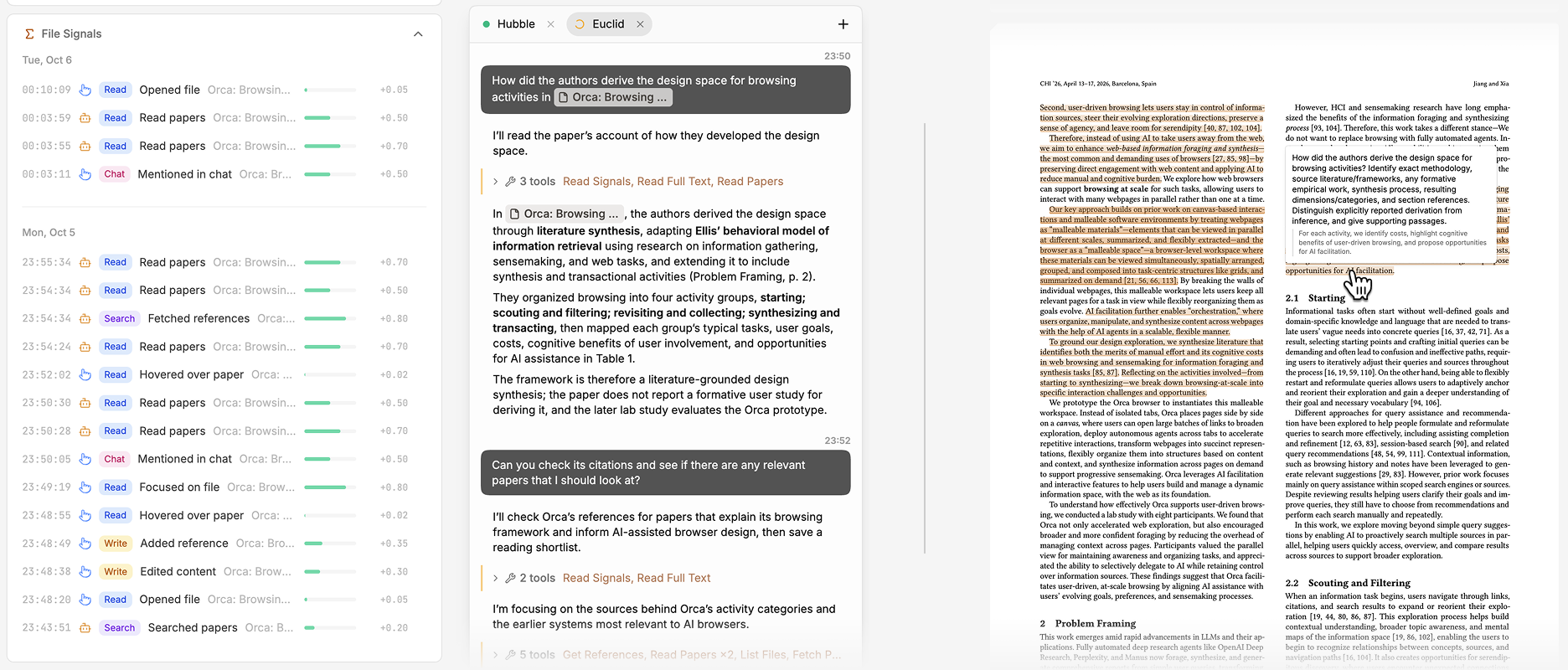}
\caption{Left: at the bottom of each file, \sys lists activities from the user and agents. Here, the agent first fetched this \texttt{.paper} file by searching. The user then mentioned it in chat, first asking a detailed question, then asking the agent to compare it with its citations. Right: the PDF view highlights passages matched to quotes returned by the agent's paper-reading tool, connecting its questions and answers to source evidence for inspection.}
\Description{Three panels show a file's chronological signal history, the user's follow-up questions in chat, and passages in the PDF highlighted to match evidence quotes returned by the agent's paper-reading tool.}
\label{fig:signals}
\label{fig:reader}
\end{figure*}

\section{Unfolding Bidirectional Awareness}

With shared artifacts grounding the collaboration (\textbf{DG1}), \sys unfolds awareness across these artifacts over time.
Specifically, \mbox{\sys} tracks both parties' interactions with the workspace over time as action ``signals'' (\mbox{\textbf{DG2}}). These signals are visualized within the file tree and documents for users and made available through query tools for agents, helping each party follow the other's work as they read, organize, and synthesize (\mbox{\textbf{DG3}}).

\subsection{Tracking Engagement Over Time (DG2)}
\label{sec:tracking}

To track how users and agents engage with artifacts over time, \mbox{\sys} records their actions on each file. Each signal includes when the action occurred, who performed it (user or agent), its category (search, read, write, chat, or organize), the specific action, the affected files and papers, and a strength value in \mbox{$[-1,1]$}. Together, these signals record both parties' engagement with the documents.

Signal strengths and denoising mechanisms are heuristic design choices informed by interaction-trace work, iterative prototyping, and pilot studies, similar to strategies used in~\cite{chang2023citesee}. Based on this exploratory work, we set the following parameters: briefly hovering over a paper or opening a file produces a weak signal (0.02 and 0.05, respectively), while sustained focus or explicitly starring a paper produces a stronger one (0.8). Agent actions follow the same approach: a search assigns a weak signal to each result (0.2), while reading metadata (0.5) or retrieving a paper's full text (0.7) assigns a stronger signal to that paper. Aggregating these strengths over time lets users and agents compare engagement across documents. Appendix~\mbox{\ref{app:strengths}} details all weights and their aggregation.

Several additional heuristics are employed to reduce incidental activity and noise. Repeated opens or hovers on the same target are recorded at most once every 30 seconds. A focus signal is recorded once after a file has remained visible and active for 30 seconds; timing pauses when the browser is hidden or loses focus. Consecutive edits and reference changes are grouped when editing pauses or the session ends. Each recorded action retains its timestamp, so users and agents can also inspect what happened and in what order.

As signals aim to help users and agents follow each other's work (\mbox{\textbf{DG2}}), these designs help make patterns of engagement visible, rather than reproduce every interaction in exact detail.

\subsection{Surfacing Awareness to Users with Localized Visualizations (DG3)}
\label{sec:visualizations}

As motivated in \textbf{DG3}, prior work on computational wear showed that embedding activity traces directly on artifacts enables awareness without separate monitoring~\cite{hill1992wear,gutwin1998design}.

\sys applies this principle by embedding awareness visualizations within the existing artifacts in the workspace, rather than in a separate dashboard:

\subsubsection*{File Tree Attention Bars}

The sidebar file tree displays compact horizontal bar indicators next to each file name (Figure~\ref{fig:overview}).
Each file shows two bars---user (blue) and agent (orange)---whose lengths reflect each party's recorded activity on that file, scaled separately for the user and agent across the visible files. Users can scan the file tree to see which papers the agent has focused on and how their own engagement is distributed across the collection. They can then open a file to inspect the specific actions in its history. This design draws on computational wear~\cite{hill1992wear}: documents accumulate visible traces of use as the review progresses.

\subsubsection*{Per-File Signal History}

At the bottom of each file, a signal history section shows the interaction history localized to that specific file---which actor performed which actions and when (Figure~\ref{fig:signals}). For example, when opening a \report file, the user can immediately see when the agent last edited it and what kinds of signals (write, read) have been recorded. This provides a fine-grained, chronological view of how both parties have engaged with a particular artifact over the course of the review.

\subsubsection*{PDF Read Wear}

When users open a paper in the built-in PDF reader, \mbox{\sys} highlights passages that match the evidence quotes returned by the agent's full-text exploration tool (Figure~\mbox{\ref{fig:reader}}). Users can read these passages in context and check whether they support the agent's answer. To locate the quotes in the PDF, the system fuzzy-matches them to sentence-level bounding boxes extracted by a Grobid-based parser~\cite{lo2023papermage,lo2024semantic}. This makes the agent's use of evidence visible within the paper the user is reading.

\subsubsection*{Inline Reference Chips}

In both \report files and user notes in \paper files, inline paper references are rendered as clickable chips. Each chip includes the user and agent engagement bars for the referenced paper. While reading a synthesis, users can see which cited papers the agent has focused on and which they have engaged with themselves, without returning to the file tree. Clicking a reference opens the paper for closer inspection.

\subsection{Surfacing Awareness to the Agent (DG3)}
\label{sec:agentic-retrieval}

The agent also needs access to the user's accumulated engagement to align its effort with the user's evolving interests.
Beyond the real-time editor context, the agent can query the signal history through a dedicated \texttt{read\_signals} tool with four modes for varied needs and targets:
\underline{(1)}~retrieving papers the user has read most recently;
\underline{(2)}~retrieving ``top papers'' within a given timeframe---those the user has repeatedly revisited, noted, or written about;
\underline{(3)}~retrieving signals for papers in a given artifact (e.g., which cited papers in a report the user has or has not yet engaged with); and
\underline{(4)}~retrieving chronological signal events to explore past action patterns, with filters for actor, action category, and time window.

\sys also generates proactive task suggestions above the chat input (Figure~\ref{fig:suggestion}). As the workspace grows, it may become harder for the user to keep track of what to do next or where the agent could help. Because \sys tracks both parties' engagement over time, it can surface timely suggestions grounded in the user's recent activity---for instance, prompting the user to incorporate notes they just took into an ongoing report, or flagging a highly cited paper they have not yet read.

\section{Example Workflow with \sys}
\label{sec:scenario}

Anna is a PhD student conducting a literature review on \emph{explainable AI for clinical decision support}. She creates a new project in \sys and uploads an early draft of her introduction as a \texttt{draft.report} file, which outlines three themes---\emph{AI explanation techniques}, \emph{clinician trust in AI recommendations}, and \emph{regulatory requirements in healthcare}---each listing a few seed papers she has read.

Anna opens the chat panel and asks the agent: ``Read my introduction and seed papers in \report[draft], then search for additional papers on each of the three themes.'' The agent reads the report, runs parallel searches, and creates four collections---\papers[existing-references], \papers[interpretability],\\ \papers[clinician-trust], and \papers[regulation]---each populated with corresponding \paper files from search results (\textbf{DG1}).
Browsing through the file tree, Anna notices that some papers whose relevance was unclear from their titles alone show increased agent attention in the sidebar (\raisebox{-2pt}[0pt][0pt]{\includegraphics[height=1.75\fontcharht\font`\B]{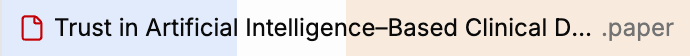}}), and opening those files reveals the detailed questions the agent asked to verify their relevance (\textbf{DG3}). She also notices that the agent moved a few papers to an \emph{Archived} folder and can inspect each to see why the agent decided they were irrelevant.

Anna finds the collections generally reasonable and trims them to focus her review. In the Interpretability section, she removes a few broad survey papers that are too general, then stars a paper on ``feature-based explanations'' that she finds particularly relevant, writing a note: ``contradicts the claim in [seed paper] that feature weights are uninterpretable---we need to compare them.'' Meanwhile, \sys records these interactions as several localized attention signals (\textbf{DG2}): opening \paper files, reading PDFs, writing notes, starring, and removing references.
The file tree's attention bars update to reflect where Anna has spent her effort.

Satisfied with the collections, Anna asks the agent to synthesize the findings. The agent knows which papers Anna has engaged with most---the ones she starred, wrote notes on, or spent time reading---and uses these signals to weight its synthesis accordingly (\textbf{DG2}, \textbf{DG3}): highly engaged papers are summarized with more details with her notes, while removed papers are excluded.
The changes are immediately reflected in the file tree, and the inline references in the report are appended with compact bar charts showing the efforts (\raisebox{-1pt}[0pt][0pt]{{\includegraphics[height=1.5\fontcharht\font`\B]{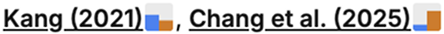}}}).\footnote{The bars show that the agent spent some effort on both papers while the user has not examined the second.} She clicks these references to navigate to \paper files and verify the work.

Later, Anna focuses the chat input, and four task suggestions appear above it, generated from her recent workspace activity and conversation (\textbf{DG3}). One references two specific papers she just read: ``Write down what makes the feature-based explanation approach in [paper a] different from the example-based methods in [paper b].'' Another suggests: ``Incorporate your recent notes on [paper c] into [report].'' She clicks the second suggestion; the agent reads the paper's full text for additional context and revises the report. After she spends time reading the Regulation section, the next round of suggestions shifts accordingly: ``List the regulatory requirements from [paper d] that most affect clinical deployment.'' These concrete next steps help her continue her exploration and complete her literature review and report.

\section{Evaluation}
\label{sec:evaluation}

We conducted two complementary studies: a lab study ($N=12$) and a longitudinal study ($N=6$). Together, they explored how artifact-centered mutual awareness supports literature review within work sessions and across repeated use. We explored these questions:

\begin{enumerate}[leftmargin=2.5em, label=\bfseries RQ\arabic*]
    \item How do participants use agent traces to inspect evidence and decide where to direct their own or the agent's attention?
    \item How do agents use researchers' workspace activity to adapt their assistance beyond explicit chat instructions?
    \item How do researchers use shared artifacts and awareness visualizations to organize reviews and coordinate multiple agents?
\end{enumerate}

\subsection{Lab Study}
\label{sec:lab-study}

\subsubsection{Participants}

We recruited 12 graduate students (P1--P12; 10 male, 2 female; aged 23--31) across multiple research universities. Participants' research fields included computer science (ML, HCI, and NLP), cognitive science, and neuroscience. We recruited participants who already used coding agents (e.g., Claude Code) frequently (7 daily, 5 a few times a week).
We selected frequent coding-agent users to reduce onboarding time and enable informed comparisons with their existing research workflows and experiences using agents. Each participant received a \$40 Amazon gift card for their time.

\subsubsection{Study Procedure}

All sessions were conducted remotely via Zoom and lasted around 60 minutes. The study included the following phases:

\paragraph{System Walkthrough (3 minutes).} Participants received a brief introduction to the research goals and a guided tour of \sys's interface. No further UI guidance was given unless requested.

\paragraph{Task 1: Understanding \sys's Agentic Behaviors by Recreating a Related Work Section (15 minutes).} Participants uploaded a related work section (\texttt{.tex} and \texttt{.bib} files) from their own paper and asked the agent to recreate it as a \report file, for which the agent searched for and retrieved the referenced papers. They then issued 1--2 follow-up prompts to expand or update the section with the latest literature.

\paragraph{Task 2: Open Exploration (30 minutes).} Participants chose a research topic they wanted to survey and were already knowledgeable about (to enable effective evaluation of agents' work). Depending on their interests, they could either continue expanding their existing related work sections or work on a different topic. They worked with the agent to discover relevant papers, organize them, and begin drafting a short synthesis.

\paragraph{Questionnaire and Interview (15 minutes).} Participants completed a five-point Likert-scale questionnaire and a semi-structured interview comparing their experience with their existing workflows and AI tools (e.g., deep research).

\paragraph{Analysis.}
Sessions were recorded and transcribed. The first author coded transcripts and observed behaviors and generated themes using both top-down and bottom-up approaches to understand how users used and assessed \sys, as well as its strengths and limitations~\cite{clarke2017thematic}.

\subsection{Findings}

In general, participants expressed strong perceived value in \sys's workflow and the artifacts they created during the study, with most (10 out of 12) asking, unprompted, to continue using \sys without compensation or to export the reports generated during the session.
Participants accumulated an average of 181 papers and sent 11 messages per session. Each request could involve several minutes of agent work, allowing us to observe repeated cycles of searching, reading, and revising.

\subsubsection{\textbf{Awareness visualizations help users inspect and steer agent work (RQ1)}}
Ten participants (all except P9--10) described a clearer understanding of agent activity relative to their previous experiences. They could see which papers the agent had worked with and inspect the evidence behind a synthesis, helping them choose what to read themselves and what to ask the agent next.

P4 used the \mbox{\explorations} (Figure~\mbox{\ref{fig:paper}}; 4) and PDF highlights (Figure~\mbox{\ref{fig:signals}}; right) to jump to source passages and verify claims about statistical analyses across papers. This connected the report to material the participant could check directly. P12 similarly inspected the source trail when checking a synthesis and appreciated access to the process \emph{``even when the final result was wrong.''}
For these participants, the workspace made verification part of reading and reviewing the agent's work. P12 also valued seeing a familiar sequence of research actions, describing the agent as \emph{``grouping them by topic, and then answering my questions.''}

Participants also found that visualizing agents' activities within the workspace where they were doing knowledge work was much more natural and easier to comprehend than examining those activities in the lengthy reasoning and tool-calling traces in the conversation panel (all except P7, P10--11). P8 contrasted this with their experience working with coding agents in IDEs, where they were \emph{``too lazy to check the conversation history [and] just kind of look at the final markdown [without understanding its process]''}---whereas signals in the file tree and inline in documents gave them real-time process awareness as a byproduct of the reading and writing they would do regardless.

Participants also used the same activity cues to choose different divisions of work.
For example, P9 used agent activity to budget where to focus their effort, always starting with papers that had the largest engagement bars because they believed these were \emph{``slightly more important papers [as agents] read more and mentioned more in the summary [report].''}
P1, on the other hand, focused on verifying the agents' coverage and attended to papers with smaller agent bars to \emph{``see [if they] are relevant as well.''}
The same display accommodated complementary strategies: P9 followed the agent's emphasis, while P1 explored sources receiving less attention. Finally, P8 found that real-time awareness allowed them to \emph{``intervene earlier''} and redirect the agent toward their intended focus while work was still underway.
Specifically, P8 stopped the agent after noticing that papers in the ``agent tool use'' subfolder---the aspect they cared about most---had been barely ``touched'' by the agent after a broad paper search. They then issued a more targeted follow-up prompt to redirect the retrieval and synthesis.

\subsubsection{\textbf{Agents respond to researchers' interests through workspace activity (RQ2)}}
Most participants noticed that agents responded not only to explicit chat instructions but also to implicit signals accumulated through their workspace activity (all except P4, P7, P8, P10).
P1 annotated the synthesis report with inline notes identifying gaps in the terms and how they might structure the literature differently, then moved on to reading other papers. When they later prompted the agent to find more papers, the agent---without being explicitly directed to---picked up on the edit signals, first reading the annotated report to capture all the notes before proceeding to a targeted search.
After browsing several related papers earlier in the session, P11 was surprised when the agent, while revising a report, surfaced a section on a topic they had not requested but found highly relevant. They noted this was \emph{``not what [they] asked for, but the agent gave [them], and it's very helpful,''} and that this was the moment the agent shifted from feeling like a search tool to a \emph{``research collaborator.''}
Similarly, when P6 posed an exploratory question about their ongoing project, the agent returned threads that precisely matched their active thinking and identified a paper they had shared with collaborators the previous week---\emph{``with almost no context that I provided explicitly.''} P6 attributed this to the integrated workspace eliminating the need to \emph{``reformulate a query and re-provide all the context.''}
These examples show how agents could learn about researchers' interests from their work beyond chat. Researchers could therefore continue reading and annotating documents while allowing agents to draw on this work without a separate explanation.

Proactive prompt suggestions accounted for approximately 11.9\% of user messages. Participants entered most requests directly and may have had clear next steps in these goal-directed sessions.

\subsubsection{\textbf{Visualizing their own activity helps researchers recognize their evolving focus (RQ1, RQ2)}}

Seeing their own engagement also helped users reflect on how their research interests were developing.

Several participants found that seeing their own engagement traces surfaced behavioral patterns they had not consciously formed (P1--2, P4, P6).
P6 noted that the system helped them \emph{``uncover patterns that [they] could not recognize [themselves],''} such as repeatedly returning to a paper across multiple reports---a pattern that eventually led them to star it.
P4 was initially surprised to have spent significant time on a paper after noticing a large blue bar. However, they then realized that it was a highly relevant paper they had been constantly drawn to.
P1 valued that their trace \emph{``naturally shows which of the files [they]'ve been looking at more''}---not only as input for the agent, but also as \emph{``a reflection of [their] own evolving focus.''}

Similarly, agents leveraged signals from other agents for coordination, which we discuss in detail in Section~\ref{sec:multi-agent}.

\subsubsection{\textbf{Researchers adapt awareness visualizations to their review strategies (RQ3)}}
We found that participants developed distinct strategies for interpreting the awareness visualizations to suit their research workflows. P6 structured a workspace with multiple reports to frame the same body of literature from different angles and themes---\emph{group conversation}, \emph{proactive assistance}, and \emph{knowledge retrieval}. The accumulated signal bars across papers then revealed distinguishable differences as different papers had been read and referenced by the agents during the syntheses, revealing core papers that were \emph{``more relevant to [them] in all aspects.''}
P6 also asked the agent to explore papers with small engagement bars in relation to a research question. The bars thus helped them identify papers that appeared across different themes and direct further work toward papers the agent had explored less.

Participants also used the file tree to check how the agent's activity was distributed across the papers cited in a report. When P12 spotted that a report over-relied on a single source, they moved to the file tree and confirmed that the paper had a correspondingly large orange bar.
Similarly, P1 found the effort visualization \emph{``helped them understand which papers the AI didn't choose [and told them] either AI intentionally didn't look at these because maybe it wasn't relevant, or it just missed it.''}

\begin{figure}
\centering
\includegraphics[width=\columnwidth]{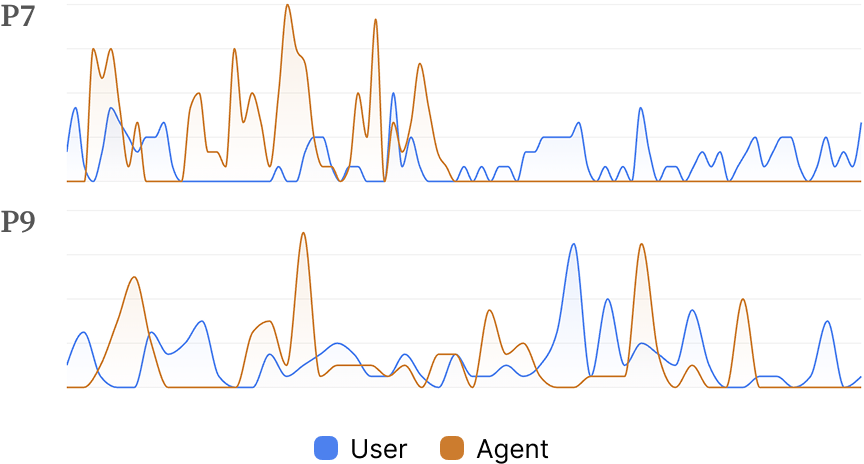}
\caption{Two literature review workflows in \mbox{\sys}. P7 first worked with agents to gather papers and develop syntheses, then focused on reading. P9 interleaved searching, synthesis, organization, and reading throughout the session.}
\Description{Two stacked activity plots compare sessions P7 and P9, with user activity in blue and agent activity in orange. P7 concentrates agent work earlier and user activity later, while P9 interleaves user and agent activity.}
\label{fig:diff-workflows}
\end{figure}

\subsubsection{\textbf{Shared artifacts support flexible literature review workflows (RQ3)}}

Most participants (all except P4, P7, and P10--11) gravitated toward reports as their primary reading and navigation surface, drilling into individual papers through inline citation links only when they needed to verify a specific claim or further explore the specific work. P3 appreciated that reports persisted as files rather than appearing ephemerally in chat, making them easy to revisit. P9 treated reports as the primary navigational hub rather than just summaries to read, noting they \emph{``would have been fine if [papers] didn't even [appear] in the file tree''} as long as they could click through from the report.

Participants used editable reports as working documents and flexible organizers. P1 added inline annotations and notes to steer the agent and valued organizing papers through themes and synthesis. The agent let them develop this organization \emph{``conversationally''} without directly manipulating the files.
P8 asked the agent to generate a comparison table of statistical analyses across papers, then used the inline links to jump to source passages and verify each entry---turning the report into an interactive audit trail.

We plotted the signals accumulated throughout the sessions as activity traces and identified diverse workflow styles echoing these differences (Figure~\ref{fig:diff-workflows})---P7 delegated substantial tasks to agents to \emph{build up the workspace} with searching and organizing before reading the accumulated materials, whereas P9 interleaved agent tasks with their own reading and editing throughout. Activity traces from all lab sessions can be found in Appendix~\mbox{\ref{app:traces}}.

\subsubsection{Study Scope and Limitations}
\label{sec:study-limitations}

This study examined how mutual awareness developed during a continuous, complete work session. We explored how participants used \mbox{\sys} and what they found useful. Their comparisons with prior workflows draw on their own experiences without a baseline or ablation.

All participants were graduate researchers and frequent coding-agent users. This sample enabled informed use with brief onboarding; broader recruitment would extend the findings to novice users and other research roles. Follow-up studies could compare matched workspaces with and without traces, separately vary visualizations and agent retrieval, and examine source coverage, claim verification, and coordination across repeated sessions. Longer deployments would also show how accumulated histories and prompt suggestions support changing research questions over weeks or months.

\subsection{Longitudinal Study}
\label{sec:longitudinal-study}

Since our design tracks, accumulates, and surfaces interaction traces of both users and agents in a workspace over time, we additionally conducted a longitudinal field deployment study to assess whether our design can scale beyond a controlled, monitored lab setting.

Six researchers (LP1--LP6; 3 male, 3 female; aged 20--27; 2 undergraduate and 4 graduate students) used \mbox{\sys} for one week on their own research tasks. After a brief onboarding with the same walkthrough as in the lab study, they were asked to spend approximately 30 minutes working in the system each day and were free to choose how to use it for their work. Participants averaged 2.8 hours of use, 215 accumulated papers, and 52 messages that week.

Three participants were frequent users of coding agents (2 daily, 1 a few times a week), while the others used them at least a few times a month. We did not observe clear differences between these groups in their usage patterns or learning curves with \mbox{\sys}. Each participant received a \$250 prepaid Visa card for their time.

We collected workspace usage logs and conducted an exit interview with each participant at the end of the week. The first author analyzed the interview transcripts and usage logs using the same process as in the lab study, comparing themes across both studies.

\subsubsection{Findings}
In general, the system's design held up during prolonged use, and the themes we found were consistent with those of the monitored lab study. This suggests that \sys was able to effectively visualize interaction traces accumulated over a much longer period for its users, and that the agents were able to effectively access them to support their work.
For example, LP1 described how agents clearly followed their \emph{``analytical intent''} through the notes and syntheses the participant had created in the workspace, without requiring them to restate that intent explicitly in chat. LP4 noted that new agents could pick up earlier agents' work \emph{``right from where they left.''}

These findings show how the shared history supported both implicit steering and continuity between the user and agents. The recurrence of these patterns supports thematic saturation within the scope of our exploratory evaluation and gives us confidence that the findings extend to continued use across sessions~\cite{guest2006saturation}.

\section{Discussion and Future Work}

Embedding action traces within documents as in-place visualizations allowed participants to inspect agent work while reading and writing. P4 followed PDF highlights to verify claims, while P8 compared engagement across a subfolder to redirect the agent. These use patterns suggest that \mbox{\sys} can support awareness at several levels: passages for checking evidence, papers for following actions, and collections for understanding how work is distributed across a topic.
Reflecting on these findings, we discuss how the awareness approach explored here can be extended in the following directions.

\subsection{Unfolding Awareness for Research Support beyond Literature Review}
\label{sec:contextual-awareness}

We developed \sys to support literature review, typically the first step in conducting research.
However, the engagement signals accumulated here---which papers a researcher starred, archived, annotated, or returned to repeatedly---encode information about their evolving understanding and interests that extend well beyond the review task.
As researchers move toward writing, agents with access to this history could better understand researchers' positioning within the field, such as which gaps they have identified and which threads they have deprioritized.
This suggests that awareness scaffolding for literature review could naturally support downstream activities such as framing contributions and drafting related work, where knowledge of \emph{how} a researcher arrived at their understanding is as valuable as the understanding itself~\cite{engestrom2014learning,kuutti1996activity}.

During iterative prototyping, we found that individual actions could be noisy cues for understanding a researcher's intent. A paper left open might reflect distraction, while repeated visits might reflect confusion. We therefore focused on capturing patterns that help both collaborators find and inspect relevant work. Cumulative bars show where activity has accumulated, while each paper's timestamped history and notes provide context for interpreting it. Users can then identify agent work worth checking, and agents can locate material worth following up on, without requiring the system to infer the purpose of every interaction.

\subsection{Unfolding Awareness for Other Domain-Specific Agentic Workflows}
\label{sec:trace-interpretation}

We envision that the approach underlying \mbox{\sys}---representing a workflow through shared, structured artifacts, tracking both users' and agents' engagement over time, and making those histories visible in the workspace and available for agents to retrieve---could also support other domain-specific workflows.
For example, in legal due diligence, the analogous artifacts might be individual contracts, clause collections grouped by risk category, and assessment briefs.
In qualitative research, they could be interview transcripts, codebooks, and thematic memos, where revisiting excerpts or merging codes could inform which themes the agent surfaces next.

An attorney could follow the activity on a contract to inspect which clauses support an assessment. Our choice to connect traces to questions and source passages lets users move from following progress to examining how a result was reached. Familiar steps such as searching and organizing may build confidence in the process, and inspecting the underlying evidence helps users decide whether to rely on a particular claim. Evaluations that vary output accuracy could further examine how evidence inspection shapes users' confidence and reliance~\cite{vereschak2021trust}.

Recording selected actions gives each signal an explicit connection to an artifact, making it directly usable for retrieval and visualization (Section~\mbox{\ref{sec:tracking}}). Complete session replays or screen recordings could retain additional context, with models identifying the relevant actions and artifacts from those recordings. As these capabilities develop, the choice of capture method may change; shared artifacts and localized visualizations would still provide a way for users and agents to inspect and coordinate their work.

The HCI community is well positioned to collaborate with domain experts to identify the artifacts, awareness signals, and workspace structures that best support agentic workflows in their respective fields.

\subsection{Unfolding Awareness in Multi-Agent~Collaboration}
\label{sec:multi-agent}

All participants created multiple agents---from P12's two agents, each focused on a distinct subtopic, to P2's seven agents, each \emph{``pursuing an intention [and] thought stream of [them].''}
Participants adopted varied strategies for working with many agents: P4 created separate workspace folders to prevent overlap, while P7 used three agents to cover distinct subtopics within a single survey.

During the study, P10 asked a second agent to extend work begun by the first. The new agent queried the accumulated signals to identify papers the previous agent had searched, organized, or queried, and began complementary work. This suggests that the same artifact-localized traces can serve as shared workspace state for both human-agent and \emph{agent-agent} collaboration: they accumulate as work proceeds and are available to every collaborator.

Users also wanted to distinguish individual agents' contributions: P2 asked about \emph{``which agent was responsible for this edit,''} and P4 wanted to know each agent's live activity.
\sys's file tree shows agents' collective efforts, while individual attribution becomes useful when assigning follow-up tasks or tracing an edit. Moving between these levels of detail would support shared workspaces with multiple users and agents~\cite{Lehmann2025CollaborativeDE}, extending workspace awareness~\cite{gutwin2002descriptive,gutwin1996workspace,gutwin1998design} to how agents divide and continue work.

\subsection{Unfolding Awareness for Proactive Agent~Assistance}
\label{sec:proactive-awareness}

The engagement history also helps agents suggest possible next steps. In \mbox{\sys}, suggestions appear beside the chat input, allowing researchers to consider them when formulating a request. This placement gives researchers control over when to take up agent initiative. More assertive interventions could bring overlooked work to their attention, but would require choosing when to interrupt~\cite{chen2025ontrack}. Building on the patterns observed in both studies, future comparisons could examine how the timing and prominence of suggestions affect their use for different research goals and activities.

Prior work on creating user models from activity traces suggests how this assistance could develop across sessions~\cite{shaikh2025gum}. The shared history in \sys could support models that relate a researcher's recent activity to interests expressed earlier. For example, an agent might suggest comparing a newly collected paper with one previously annotated as addressing a key research question. Keeping recommendations connected to the artifacts would allow researchers to inspect the context behind suggestions.

\section{Conclusion}

We introduced \mbox{\sys}, a shared workspace for AI-assisted literature review that unfolds mutual awareness between users and agents through structured artifacts, bidirectional engagement tracking, and localized visualizations. Our lab and longitudinal studies with eighteen researchers showed how embedding awareness signals within the documents and file structures where users already work helped participants understand, inspect, and steer agent work. These visualizations also helped users recognize their own latent interests, while the shared signal history enabled agents to coordinate implicitly. These findings extend mutual-awareness principles from human-human collaboration to human-agent workspaces and offer a reusable design approach to collaboration across domains.

\begin{acks}
Thanks to the anonymous reviewers for their helpful comments.
\end{acks}

\bibliographystyle{ACM-Reference-Format}
\bibliography{references}

\appendix

\section{Implementation Details}
\label{sec:implementation}

\sys is a web application. The frontend uses React, and the backend uses Elysia with Bun. Both are written in TypeScript, totaling $\sim$66,000 lines.
We use PostgreSQL for data storage and the Yjs CRDT library for real-time collaborative editing.
Agent edits are routed through the same Yjs collaboration layer as users' edits, so that users can see them in real time, and all edits are recorded in the Yjs revision history with attribution at the character level.

The \sys agent is built with the Pi agent SDK\footnote{\url{https://github.com/badlogic/pi-mono/}}.
The agent's literature review tools are built on Semantic Scholar APIs\footnote{\url{https://www.semanticscholar.org/product/api}} for paper metadata and full-text retrieval.
All the latest agentic features are available, including skills (e.g., writing reports in different styles or formats) and the ability to dispatch subagents with specialized prompts (e.g., a reviewer agent after the report is created to identify issues).
The main agents use \texttt{gpt-6-astra} or \texttt{gpt-6.1-sol}, according to the user's model selection. Auxiliary processes, including search-result filtering, project-title generation, and prompt suggestions, default to \texttt{gpt-6.1-sol} for better responsiveness.

To support ``read wear'' visualization in the PDF reader, the system uses a Grobid-based PDF parser to extract sentence-level bounding boxes from papers~\cite{lo2023papermage, lo2024semantic}. When the agent reads a paper, it is prompted to include verbatim evidence quotes alongside its answers. These quotes are then fuzzy-matched to parsed sentence bounding boxes, allowing the PDF reader to highlight the specific passages the agent drew on.

\section{Action Signal Strengths}
\label{app:strengths}

Table~\mbox{\ref{tab:signals}} lists heuristic weights for summarizing recorded actions. These summarize interaction rather than measuring attention or intent. Positive values add to a signed engagement score, while negative values encode actions such as removing or downvoting.

For an artifact and actor, the activity bar sums positive weights of matching events. It applies square-root normalization against that actor's maximum among displayed files, with a minimum visible proportion of 0.1 for nonzero values. User and agent bars therefore show relative activity within each actor's history.

\begin{table}[t]
\centering
\small
\caption{Action signal weights.}
\label{tab:signals}
\begin{tabular}{@{}llr@{}}
\toprule
\textbf{Actor} & \textbf{Action} & \textbf{Strength} \\
\midrule
User & Hover over paper card & 0.02 \\
 & Open file & 0.05 \\
 & Active file focus (30 s) & 0.80 \\
 & Highlight PDF passage & 0.50 \\
 & Add / edit highlight note & 0.30 \\
 & Mention in chat & 0.50 \\
 & Edit content & 0.30 \\
 & Add / remove reference & \mbox{$+0.35 / -0.45$} \\
 & Update file metadata & 0.15 \\
 & Upload files & 0.20 \\
 & Create folder & 0 \\
 & Add paper to folder & 0.30 \\
 & Remove paper from folder & \mbox{$-0.50$} \\
\midrule
Agent & Search papers (per result) & 0.20 \\
 & Fetch papers & 0.50 \\
 & Fetch citations/references (source) & 0.80 \\
 & Read metadata / file / highlights & 0.50 \\
 & Full-text exploration & 0.70 \\
 & Edit file / PDF annotations & 0.50 \\
\midrule
Both & Create file & 0.50 \\
 & Move / copy files & 0.10 \\
 & Rename file / folder & 0.20 \\
 & Delete files / folders & \mbox{$-0.70$} \\
 & Star / unstar & \mbox{$+0.80 / -0.80$} \\
 & Downvote / remove downvote & \mbox{$-0.80 / +0.80$} \\
\bottomrule
\end{tabular}
\end{table}

\onecolumn
\section{Activity Traces for User Study Sessions}
\label{app:traces}

\begin{figure}[ht]
  \centering
  \includegraphics[width=\textwidth]{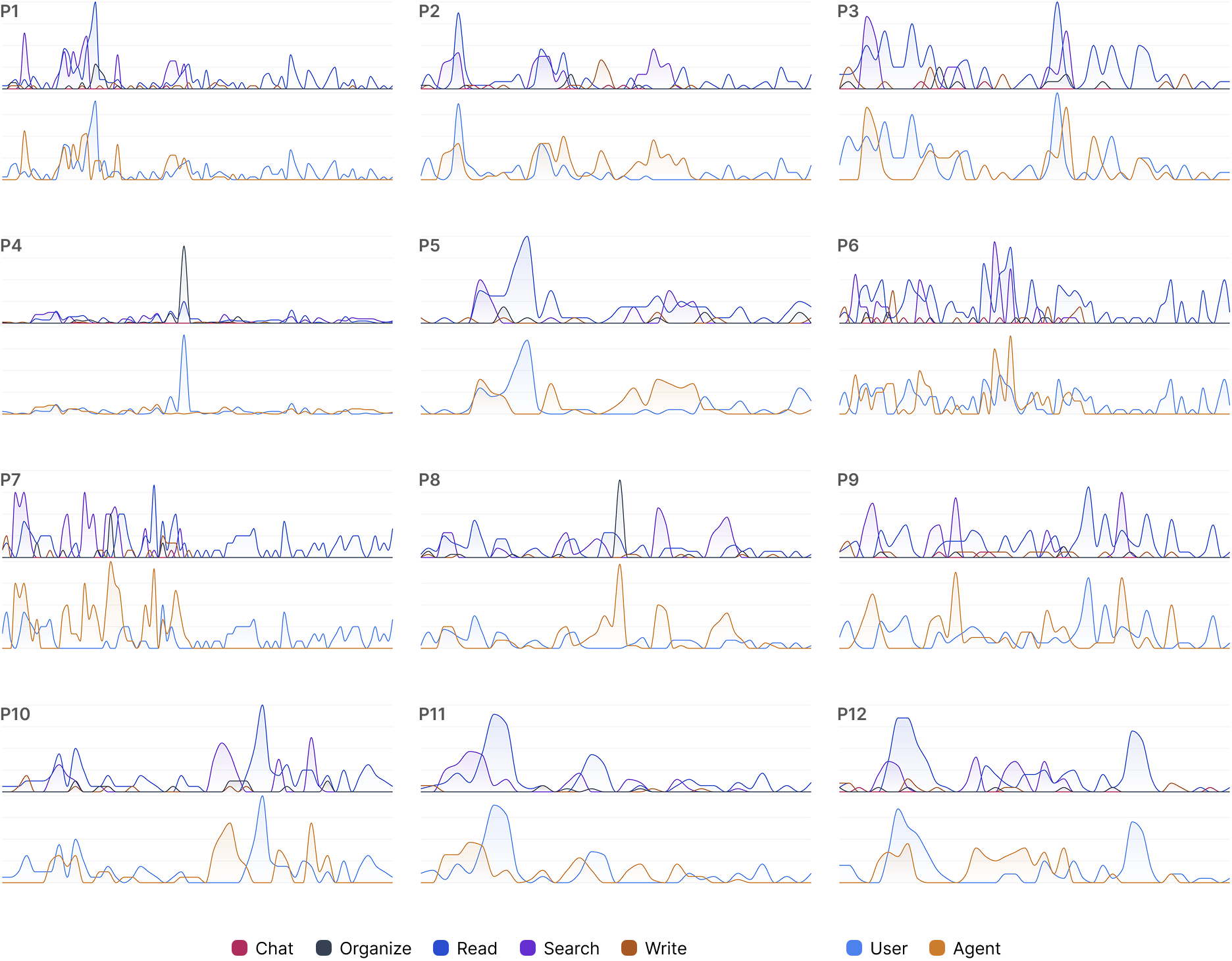}
  \caption{Activity traces for users and agents, and for different action types in user study sessions.}
  \Description{Twelve panels show activity traces for participants P1 through P12. Each panel compares user and agent activity over the session and separates recorded actions into chat, organize, read, search, and write categories.}
  \label{fig:traces}
\end{figure}

\end{document}